\documentclass{article}

    \usepackage[preprint]{neurips_2025}

\usepackage[utf8]{inputenc} 
\usepackage[T1]{fontenc}    
\usepackage{hyperref}       
\usepackage{url}            
\usepackage{booktabs}       
\usepackage{amsfonts}       
\usepackage{nicefrac}       
\usepackage{microtype}      
\usepackage{xcolor}         
\usepackage{amsmath}

\definecolor{darkgreen}{rgb}{0.0, 0.5, 0.0} 
\definecolor{lightblue}{RGB}{220,235,250}
\definecolor{lightgray}{RGB}{211,211,211}
\definecolor{lightgreen}{RGB}{230, 255, 230}
\definecolor{mygray}{gray}{0.85}  

\newcommand{\cred}[1]{\textcolor{red}{$_{#1}$}}
\newcommand{\cgreen}[1]{\textcolor{darkgreen}{$_{#1}$}}
\usepackage{multirow}
\usepackage{wrapfig}

\usepackage{graphicx}     
\usepackage[most,skins,theorems]{tcolorbox}

\newcounter{takeawaycounter}
\renewcommand{\thetakeawaycounter}{\arabic{takeawaycounter}}

\makeatletter
\@addtoreset{takeawaycounter}{nothing} 
\makeatother

\newcommand{\nexttakeawaytitle}{%
  \stepcounter{takeawaycounter}%
  Finding~\thetakeawaycounter%
}

\tcbset{
  takeawaysbox/.style={
    title={\nexttakeawaytitle}, 
    colback=lightblue!80,
    colframe=black,
    fonttitle=\bfseries\small,
    coltitle=white,
    colbacktitle=black,
    enhanced,
    attach boxed title to top left={xshift=2.5mm,yshift=-2.5mm},
    boxed title style={rounded corners, size=small, 
    colframe=black, colback=black},
    width=\linewidth,
    arc=2.5mm
  }
}

\newtcolorbox{mytakeaway}[1][]{
  takeawaysbox,
  label={takeaway:\thetakeawaycounter},
  #1
}

\usepackage{enumitem}     
\usepackage{algorithm}    
\usepackage{algorithmic}  
\usepackage{float}        
\usepackage{bm} 
\usepackage{CJKutf8}
\usepackage{subcaption}
\usepackage{colortbl}
\usepackage{arydshln}
\usepackage[textsize=tiny]{todonotes}
\definecolor{newgreen}{rgb}{0.7, 0.9, 0.7}
\definecolor{newblue}{rgb}{0.85, 0.85, 0.9}
\usepackage{amssymb} 
\usepackage{xspace}
\usepackage{cleveref}
\newcommand{\modelname}{{\bf CoRE}\xspace}
\newcommand{\model}{{\bf CoRE-Eval}\xspace}

\definecolor{tkcolor}{RGB}{224,223,255}
\definecolor{lgcolor}{RGB}{240,240,240}

\title{CoRE: Enhancing Metacognition with Label-free Self-evaluation in LRMs}

\author{%
  Haoxi Li \\
  HKUST \\
  Hong Kong, China \\
  \texttt{hligb@connect.ust.hk} \\
  \And
  Sikai Bai \\
  HKUST \\
  Hong Kong, China \\
  \texttt{sbaiae@connect.ust.hk} \\
  \AND
  Jie Zhang \\
  HKUST \\
  Hong Kong, China \\
  \texttt{csejzhang@ust.hk} \\
  \And
  Song Guo \\
  HKUST \\
  Hong Kong, China \\
  \texttt{songguo@cse.ust.hk} \\
}

\begin{document}

\maketitle

\begin{abstract}
Large reasoning models (LRMs) have demonstrated impressive capabilities in domains like mathematics and program synthesis. Despite their strong performance, LRMs often exhibit overthinking---excessive and redundant reasoning steps that introduce inefficiencies during inference. This phenomenon raises an important question for LRM self-evaluation: \textit{How can a model autonomously assess the correctness of its own reasoning trajectory without external labels?} To address this, we propose \textbf{Chain-of-Reasoning Embedding (CoRE)}, a series of hidden states in latent space to enable label-free self-evaluation on intermediate reasoning steps of LRMs, so as to enhance metacognition abilities for improved reasoning efficiency. 
By analyzing the geometric properties of the \modelname trajectories, we reveal that redundant reasoning usually presents cyclical fluctuations, which correspond to repetitive and unconscious reflection/exploration.  
Leveraging this insight, we further introduce a training-free, label-free self-evaluation framework, \model, to detect such patterns and dynamically determine whether to terminate reasoning early. 
Extensive experiments on mathematical reasoning benchmarks (\textsc{GSM8K}, \textsc{MATH-500}, and \textsc{AIME}) and across model sizes from 7B to 32B demonstrate that \textbf{CoRE-Eval} reduces chain-of-thought length by 13.7\% to 33.2\% while improving answer accuracy by around 10\%, achieving 70.0\% accuracy on the challenging \textsc{AIME} benchmark with the 32B model.

\end{abstract}

\section{Introduction}
\label{Sec/introduction}

Large Reasoning Models (LRMs) have recently emerged as powerful tools for complex reasoning, with prominent examples such as OpenAI O1~\cite{jaech2024openai} and DeepSeek R1~\cite{guo2025deepseek} demonstrating impressive capabilities in domains like mathematics and program synthesis~\cite{sun2023survey, yu2024natural, team2025kimi}. These models distinguish themselves through their capacity for test-time scaling, predominantly by producing extended Long Chain-of-Thought (Long CoT) that elaborate reasoning pathways with rich intermediate steps~\cite{team2025kimi, chen2025towards, wei2022chain, wang2022self}. While such extensive thought generation enables more nuanced reasoning, it also risks introducing inefficiencies and redundancies, a phenomenon commonly known as ``overthinking''~\cite{team2025kimi, chen2024not}. To address this, exploring self-evaluation mechanisms~\cite{huang2025survey, wang2024latent} in LRMs becomes critical to enhance the reliability and optimize the efficiency of the model's reasoning processes.

Recent investigations into the reasoning behavior of LRMs indicate that these models often exhibit an inherent awareness of the correctness of their own reasoning. This insight motivates a natural strategy for mitigating overthinking: evaluating the validity of intermediate reasoning steps to decide whether to early exit reasoning during inference. 
One of the most prominent self-evaluation methods~\cite{yang2025dynamic} employed during reasoning lies in operating within the output space, for example, evaluating the confidence to immediately generate a trial answer based on some specific tokens such as \texttt{``Wait''}, \texttt{``But''}~\cite{yang2025dynamic}. 
However, its over-reliance on heuristically pre-defined transition points may hinder their generalization abilities across diverse tasks or varied reasoning styles.
In this case, some recent studies have turned to explore the presence of ``look-ahead'' information embedded within the hidden state of LRM reasoning steps in latent space~\cite{zhang2025reasoning}. 
Although such implicit estimation methods show more potential and may offer greater interpretability, these output-free research often require correctness labels 0/1 to formulate the training dataset, e.g., for learning probing classifiers to verify the stopping points of reasoning.
Building upon this, we are motivated to
address the challenging yet crucial question: \textbf{How can we leverage solely the latent representations (hidden states) to estimate LRM reasoning redundancy without any label?}

\begin{wrapfigure}{r}{0.57\textwidth}
\vspace{-0.10in}
  \centering
  \includegraphics[width=0.55\textwidth]{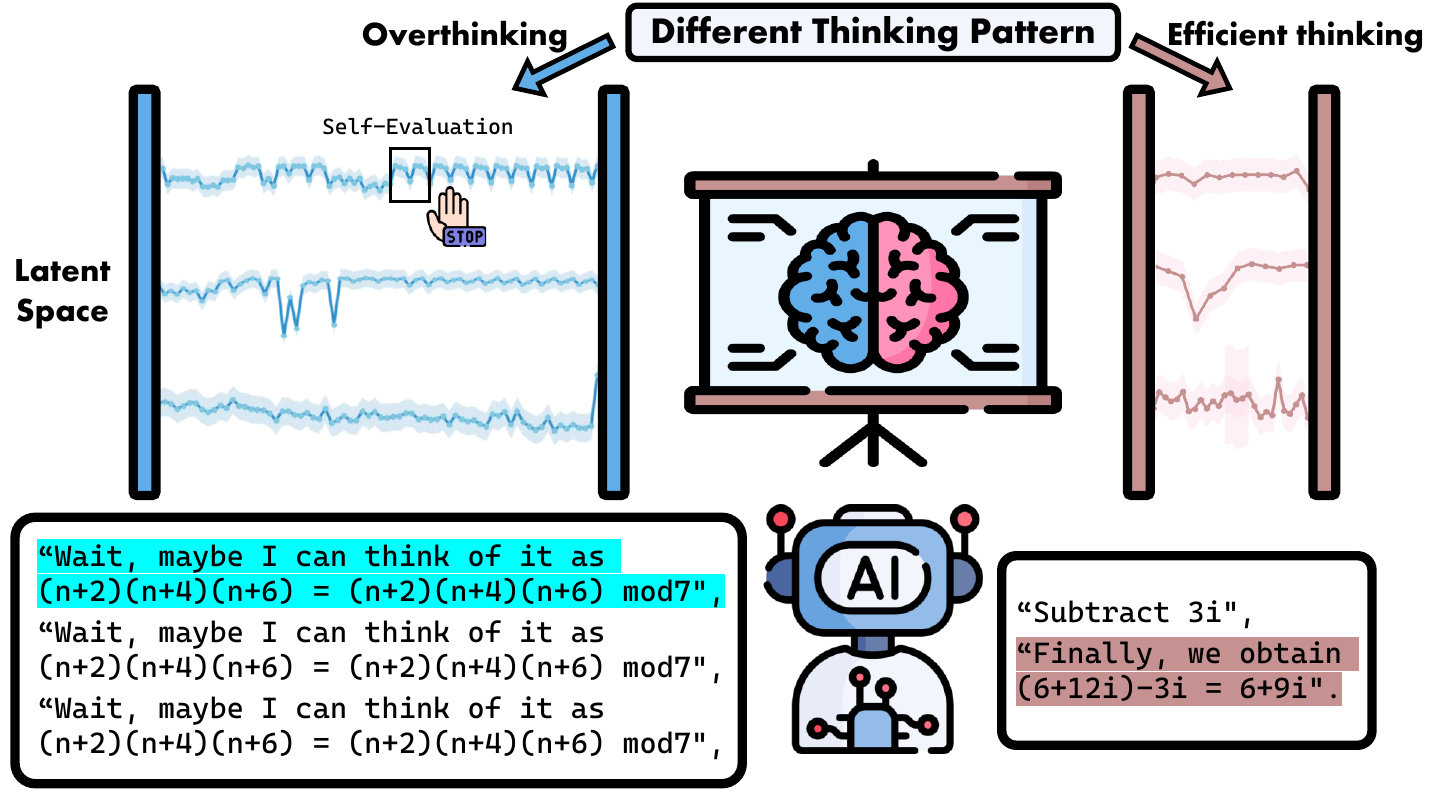}  
  \caption{Analogous to human cognition, distinct reasoning patterns yield divergent trajectories in the Chain-of-Reasoning Embedding space.}
  \label{img:first-figure}
\vspace{-0.10in}
\end{wrapfigure}

From the perspective of human intelligence, we found that effective human reasoning, often guided by principles such as satisficing~\cite{Simon1956} and adaptive metacognitive control~\cite{Fletcher2012}, tends to follow a pruned cognitive trajectory---one that terminates deliberation once a sufficiently confident and satisfactory solution is reached. In contrast, overthinking in humans manifests as a more convoluted and extended trajectory, often marked by prolonged or ruminative deliberation that burdens working memory~\cite{Deck2015} and leads to diminishing returns~\cite{Fletcher2012}. These divergent cognitive patterns---one efficiently truncated, the other prone to excessive elaboration---are similarly observed in large reasoning models (LRMs), as illustrated in Fig.~\ref{img:first-figure}. This parallel suggests that analogous dynamic signatures in LRMs latent states could provide the basis for a label-free mechanism to detect and mitigate overthinking.
%

%
Inspired by the analyses above, in this paper, we propose \model, a label-free self-evaluation framework for reasoning models. Our approach represents the sequence of step-level hidden states as a latent trajectory in embedding space, referred to as the \textbf{Chain-of-Reasoning Embedding (CoRE)}. 
To achieve a training-free self-evaluation and enhanced reasoning process, we further design a \textbf{Local Dynamics Calculation} module on \modelname through two interpretable geometric signals---magnitude and angle---to reflect different reasoning patterns, i.e., strategic reasoning, semantic redundancy, and cognitive stagnation; and a \textbf{Cyclic Redundancy Detection} module based on a composite signal and a sliding-window mechanism, to identify transition points that can be terminated early during reasoning.
Extensive experiments across standard mathematical reasoning benchmarks, including \textsc{GSM8K}~\cite{cobbe2021training}, \textsc{MATH-500}~\cite{hendrycks2021measuring}, and \textsc{AIME 2024}, demonstrate that \textbf{CoRE-Eval} consistently reduces Long CoT length by 13.7\% to 33.2\%, while also improving average accuracy by around 3.6\%. Particularly noteworthy is its performance on the DeepSeek-R1-Distill-32B model, where it achieves 70.0\% accuracy on \textsc{AIME 2024}, surpassing the original baseline by 10\% and highlighting the method's scalability and effectiveness in mitigating overthinking.

    
    
    
Our main contributions are summarized as follows:\vspace{-5pt}
\begin{itemize}[leftmargin=4ex]
    \item \textbf{Label-free self-evaluation framework:} We introduce \textbf{CoRE-Eval}, the first label-free self-evaluation mechanism for large reasoning models, grounded in geometric analysis of their latent reasoning trajectories.
    
    \item \textbf{Geometric reasoning diagnostics:} We propose \textbf{CoRE}, a Chain-of-Reasoning Embedding that captures the evolution of hidden states across reasoning steps. By decomposing trajectories into magnitude and angle signals, we uncover dynamic patterns associated with exploration, redundancy, and cognitive stagnation.
    
    \item \textbf{Cyclic redundancy detection and early-exit mechanism:} We develop an efficient algorithm to detect quasi-cyclic reasoning patterns via a composite signal and sliding-window correlation, combined with a finite-state hysteresis controller that governs stable early termination of unproductive reasoning.
    
    \item \textbf{Empirical validation across models and benchmarks:} Experiments on \textsc{GSM8K}, \textsc{MATH-500}, and \textsc{AIME 2024} show that \textbf{CoRE-Eval} reduces reasoning length by up to 35.9\% while improving answer accuracy by up to 3.6\%. On \textsc{AIME 2024}, it enables the 32B model to reach 70.0\% accuracy, a 10\% improvement over the baseline.
\end{itemize}

\section{Related Work}
\label{Sec/related_work}

\textbf{Self-Evaluation in LLMs.}
The goal of self-evaluation in LLMs is to estimate whether the LLM response to a given input question is correct or not through LLMs' own capabilities, so as to enhance the practical applicability of LLMs in real-world scenarios~\cite{wang2024latent}. 
Recent research on LLMs self-evaluation can be roughly categorized into two types: 
1) \textbf{Explicit estimation} based on well-designed prompts or specific output tokens. The former one prompts the LLM to directly validate its responses or intermediate steps based on its internal knowledge  (e.g., adding prompts
like ``\texttt{Is the proposed answer True or False?}'')
\cite{zhang2023coder,shinn2023reflexion,kadavath2022language,tian2023just,gao2024spuq}, requiring empirical design for the prompts and may lead to unstable performance. Another line of work focuses on leveraging specific semantic tokens (e.g., ``\texttt{Wait}'', ``\texttt{But}'') that indicate reasoning transition points to \cite{yang2025dynamic} to evaluate the confidence for immediately generating a trial answer. However, this paradigm may have poor generalization ability across different tasks due to the explicit reliance on critical tokens.   
2) \textbf{Implicit estimation} based on latent-space embeddings, which reveals that the latent space of LLMs is more informative in reflecting the response correctness than LLM output~\cite{duan2024llms, li2023inference,burns2023discovering} in an interpretable manner, such as by training probing classifiers to predict the correctness~\cite {su2024unsupervised,zhang2025reasoning}. However, these approaches often require correctness labels 0/1 to formulate the probing dataset, hindering its further application in open-world scenarios. This inspires us to explore a label-free self-evaluation method to further enhance the reasoning performance in LRMs.

\textbf{Test-time Efficient Reasoning.}
To alleviate the ``overthinking'' phenomenon in LRMs, many researchers have explored efficient test-time methods to make reasoning more concise and smart by optimizing the length of generated reasoning steps and eliminating unnecessary steps~\cite{sui2025stop,wang2025harnessing}. Different from post-training based efficient reasoning methods that rely on supervised fine-tuning~\cite{kang2025c3ot,xia2025tokenskip,ma2025cot,munkhbat2025self} with variable-length CoT data or reinforcement learning with well-designed length rewards~\cite{team2025kimi,luo2025o1,aggarwal2025l1,shen2025dast,cui2025stepwise}, test-time efficient reasoning methods get rid of the large amount of computational resource consumption and potential challenges in dataset construction. For instance, prompt-based approaches~\cite{xu2025chain,lee2025well,chen2024unlocking,ma2025reasoning} use varying prompts to enforce LRMs to generate concise CoT with fewer reasoning steps (e.g., setting a token budget in prompts). Other approaches focus on controlling the reasoning strategies by selecting appropriate criteria~\cite{wu2025more,aytes2025sketch,liao2025reward,ding2025dynamic}, such as using a lightweight router model to decide the optimal reasoning paradigm~\cite{aytes2025sketch} or speculative-based reasoning~\cite{liao2025reward}. 
Despite this progress, we find that all the above methods fail to adequately utilize the latent state information during inference time to enhance the cognitive ability on reasoning process. 

\textbf{Summary.}
To sum up, investigating novel self-evaluation methods in LRMs to achieve efficient test/inference-time reasoning becomes critical for us to align human intelligence. Different from existing works on efficient reasoning that either rely on manually designing some supervision signals, which inevitably require fine-tuning or a reinforcement learning process, or explore LRMs' inherent characteristics (e.g., hidden state in the latent space) to conduct self-evaluation in a plain manner, we are the first to propose a test-time training-free and label-free self-evaluation framework based on a well-designed \textbf{Chain-of-Reasoning Embedding (CoRE)}, with the goal of enhancing the metacognition of LRMs on complex problem reasoning.
\section{Definition of Chain-of-Reasoning Embedding (CoRE)}
\label{sec:trajectory_embedding}
Suppose a CoT consists of a sequence of $T$ reasoning steps $\mathcal{S} = (S_1, S_2, \dots, S_T)$, where each step $S_t$ represents a segment of natural language representing an intermediate thought. To formalize the reasoning dynamics, we define the reasoning process in a latent space as follows:

\textbf{Step Embedding.} 
Let $f: \mathcal{S} \to \mathbb{R}^d$ be the embedding function that maps each reasoning step to its latent representation. Following the definitions in \cite{zhang2025reasoning}, we define the last-layer hidden states at the last token position as its representation:
\begin{equation}
    \bm{h}_t = f(S_t) = \text{Encoder}^{(L)}(S_t)[-1] \in \mathbb{R}^d
    \label{eq:encoder}
\end{equation}
where $\text{Encoder}^{(L)}(\cdot)$ denotes the hidden states of the final transformer layer (layer $L$), The notation $[-1]$ indexes the last token's embedding in the generated step sequence, and $S_t$ is processed as an input segment with its own positional encoding.

\textbf{Chain-of-Reasoning Embedding (CoRE).}
The sequence of the step embeddings forms the \modelname, denoted as $\tau$. This trajectory captures the evolution of the LRM's internal state throughout the reasoning process:
\begin{equation}
    \tau = (\bm{h}_1, \bm{h}_2, \dots, \bm{h}_T) \in \mathbb{R}^{T \times d}.
    \label{eq:trajectory}
\end{equation}
The length $T$ of the trajectory can vary for different reasoning instances. \modelname serves as the foundation for our self-evaluation mechanism.

\section{Methodology}
\label{sec:method}
In this section, we present our self-evaluation framework, dubbed \model, for enhancing the metacognition ability of LRMs.
Specifically, the objective is to enable LRMs to autonomously detect and mitigate redundant reasoning steps, thereby improving inference efficiency without external supervision. Building upon the \modelname~(Sec.~\ref{sec:trajectory_embedding}), our method comprises two main stages:
1)~\textbf{Local Dynamics Calculation}~(Sec.~\ref{sec:state_transition_diag}), which characterizes the local dynamics of the reasoning trajectory $\tau$ using geometric measures derived from consecutive hidden states;
2) \textbf{Cyclic Redundancy Detection}~(Sec.~\ref{sec:cyclic_detection}), which analyzes these diagnostics to identify recurrent, potentially uninformative patterns in the latent space. Upon detection of such patterns, an early-exit policy is triggered (Sec.~\ref{sec:early_exit}).

\subsection{Local Dynamics Calculation on CoRE}
\label{sec:state_transition_diag}
To analyze the internal evolution of long reasoning within LRMs, we investigate the geometric properties of the latent trajectory $\tau$ from two aspects~\cite{helland2009trajectory, rintoul2015trajectory}: \textbf{magnitude} and \textbf{angle}, which can reflect the distance and direction of representational changes throughout the reasoning process.
%
We formally define the changes in magnitude and angle~\footnote{While the true angular difference is given by $\arccos(c_t^{\mathrm{ang}})$, we adopt cosine similarity $c_t^{\mathrm{ang}}$ as a proxy for computational simplicity and consistent interpretability.} between adjacent hidden states $(\bm{h}_{t}, \bm{h}_{t+1})$ for $t \in \{1, \dots, T-1\}$ as:
\begin{equation}
\label{eq:mag_delta}
\begin{aligned}
\delta_t^{\mathrm{mag}} &= \|\bm{h}_{t+1} - \bm{h}_{t}\|_2, \qquad
c_t^{\mathrm{ang}} &= \frac{\langle \bm{h}_{t}, \bm{h}_{t+1} \rangle}{\|\bm{h}_{t}\|_2 \, \|\bm{h}_{t+1}\|_2}.
\end{aligned}
\end{equation}
Here, $\delta_t^{\mathrm{mag}}$ quantifies the magnitude change using the L2-norm, while $c_t^{\mathrm{ang}}$ measures the cosine similarity between $\bm{h}_t$ and $\bm{h}_{t+1}$, indicating the directional change. The sequences $\{\delta_t^{\mathrm{mag}}\}_{t=1}^{T-1}$ and $\{c_t^{\mathrm{ang}}\}_{t=1}^{T-1}$ describe the LRM's reasoning dynamics.
%
%

\textbf{CoRE Visualization.} 
To investigate the influence of \modelname's geometric features, we visualize the angle and magnitude sequences of its latent reasoning trajectories, with a particular emphasis on Long CoT reasoning. Our analysis focuses on standard mathematical reasoning benchmarks, including \textsc{Math}, \textsc{GSM8K}, and \textsc{AIME2024}. The dataset descriptions and references are provided in Sec.~\ref{sec:exp-setup}. All visualizations are based on the R1-Distill-Qwen2.5-7B model as the backbone. A representative example is shown in Fig.~\ref{img:trajectory}, with additional cases provided in the Appendix.

\begin{figure}[t]
    \centering
    \includegraphics[width=0.97\textwidth]{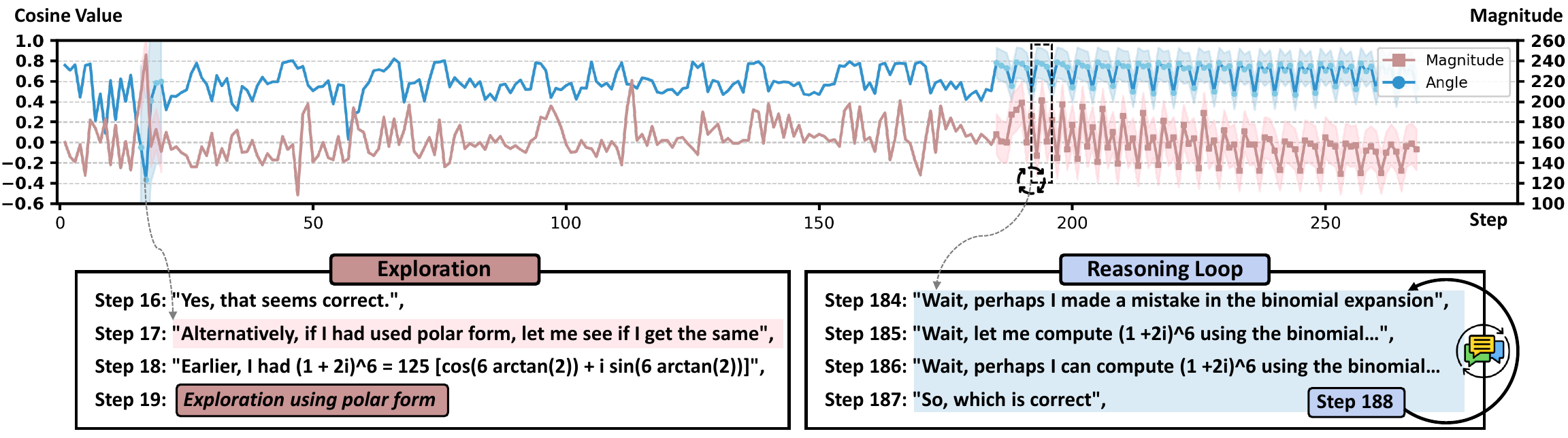}
    \caption{{\bf CoRE Visualization.} \textcolor{blue}{Blue} and \textcolor{red}{Red} Long CoT trajectories represent changes in the \textcolor{blue}{angle} and \textcolor{red}{magnitude} dimensions, respectively. Shaded regions highlight notable reasoning patterns.}
    \vspace{-0.2in}
    \label{img:trajectory}
\end{figure}

We observe sudden changes in magnitude and sharp reversals in angle (e.g., cosine similarity transitioning from positive to negative), which often align with cognitive shifts such as hypothesis revision or strategic exploration in human intelligence. For instance, at step~17, the phrase ``\texttt{Alternatively, if I had used polar form, ...}'' corresponds to prominent extrema in both magnitude and angle, indicating an exploratory reasoning pattern characteristic of Long CoT generation.

\begin{mytakeaway}
Abrupt changes in \modelname dynamics can serve as indicators of strategic exploration or conceptual revision in reasoning.
\end{mytakeaway}

In later steps, both magnitude and angle patterns exhibit quasi-periodic oscillations (e.g., Step 184-187). These trajectories correspond to repetitive verbal patterns and local semantic redundancy (e.g., repeated binomial expansion calculations). The recurrent geometric patterns suggest that \modelname can detect reasoning loops, which often signify inefficiency and overthinking in LRM outputs.

\begin{mytakeaway}
\label{finding:2}
    Quasi-periodic fluctuations in \modelname trajectories reliably signal reasoning loops and verbal redundancy, reflecting inefficient cognitive processing.
\end{mytakeaway}

Throughout the trajectory, we observe an inverse correlation between angle and magnitude: high cosine similarity between steps often coincides with small latent shifts in embedding space, and vice versa. This pattern reflects a state where the model generates semantically redundant steps without substantial cognitive progression.

\begin{mytakeaway}
    The anti-correlation between angular similarity and magnitude reveals latent stagnation zones in reasoning, enabling detection of repetition and inefficient cognitive loops.
\end{mytakeaway}

\subsection{Cyclic Redundancy Detection for Self-Evaluation}
\label{sec:cyclic_detection}

In Sec.~\ref{sec:state_transition_diag}, we quantified \modelname's geometric properties and highlighted key findings in long CoT reasoning. Building on the quasi-periodic fluctuation pattern identified in \textbf{Finding~2}, we now introduce a label-free self-evaluation algorithm, \model, that detects cyclic redundancy in reasoning trajectories and enables early exit to mitigate overthinking.

\begin{figure}[t]
    \centering
    \includegraphics[width=0.97\textwidth]{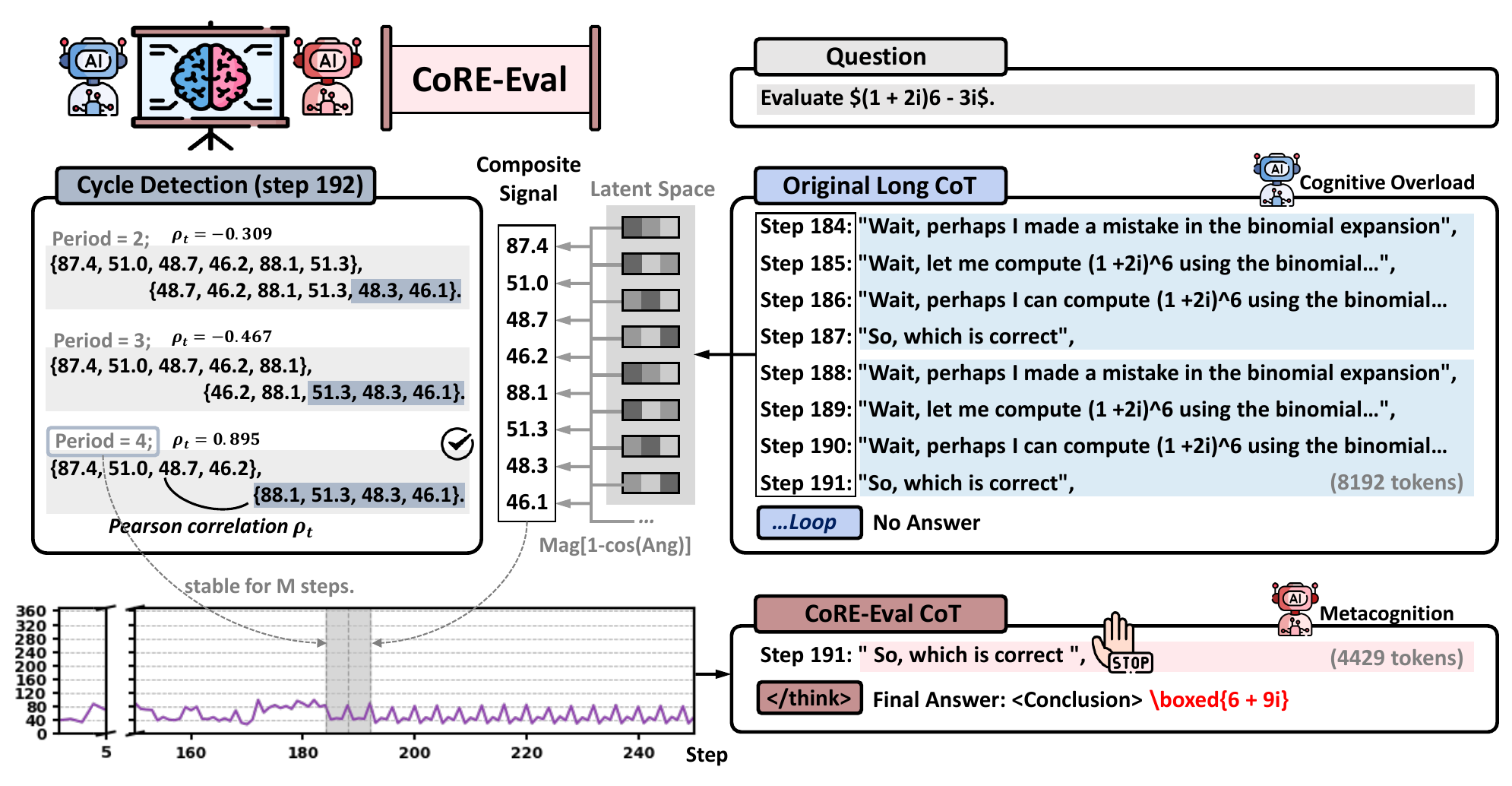}
    \caption{An overview of the Chain-of-Embedding Reasoning Self-evaluation  (\textbf{CoRE-Eval}) method.}
    \vspace{-0.2in}
    \label{img:framework}
\end{figure}

\textbf{Composite Signal Definition.}
Motivated by the inverse correlation between magnitude and angular similarity observed in \textbf{Finding~3}, we introduce a composite signal designed to capture significant deviations in the \modelname trajectory. Such composite approaches are commonly used in neural trajectory analysis~\cite{chaudhuri2019dynamics,gao2022understanding} to detect representational shifts. Specifically, we define:
\begin{equation}
\label{eq:composite_signal}
z_t = \delta_t^{\mathrm{mag}} (1 - c_t^{\mathrm{ang}}),
\end{equation}
where $\delta_t^{\mathrm{mag}}$ is the L2-norm magnitude between step $t$ and $t+1$, and $c_t^{\mathrm{ang}}$ is their cosine similarity.

This formulation ensures that $z_t$ attains higher values when the model exhibits large representational shifts accompanied by semantic divergence, capturing moments of cognitive redirection or exploration. In contrast to using either signal alone, this combined measurement is more sensitive to meaningful deviations in the model's reasoning path, facilitating the evaluation of cyclic redundancy.

\textbf{Local Periodicity via Sliding Window.}
To capture periodic redundancy in non-stationary reasoning trajectories, we employ a step-wise sliding window with a fixed size of $W$. Within each window, we compare the most recent composite signal trajectory with its temporally shifted (lagged) counterpart, evaluating their similarity across a range of candidate periods $\ell \in [1, P_{\max}]$, where $P_{\max}$ denotes the maximum period of interest. If these two sequences are highly correlated, it suggests that the trajectory is re-entering a previously visited semantic state, indicating the presence of cyclic behavior.

Specifically, at each inference step $t$, for every candidate period $\ell$, we compute the Pearson correlation~\cite{pearson1895vii}---a standard measurement of linear similarity invariant to scale---between the current step sequence of length $W-\ell$ and its lagged version. We first normalize both sequences to have zero mean and unit variance:
\begin{equation}
\hat{z}_{t-i} = \frac{z_{t-i} - \mu_t^{(\ell)}}{\sigma_t^{(\ell)}}, \quad
\hat{z}_{t-i-\ell} = \frac{z_{t-i-\ell} - \mu_{t-\ell}^{(\ell)}}{\sigma_{t-\ell}^{(\ell)}},
\end{equation}
where $\mu_t^{(\ell)}$ and $\sigma_t^{(\ell)}$ denote the mean and standard deviation of the current signal segment $\{z_{t-i}\}_{i=1}^{W-\ell}$, while $\mu_{t-\ell}^{(\ell)}$ and $\sigma_{t-\ell}^{(\ell)}$ are those of the lagged sequence $\{z_{t-i-\ell}\}_{i=1}^{W-\ell}$.

Based on the normalized sequences, we compute the Pearson correlation coefficient as:
\begin{equation}
\label{eq:zscore_corr}
r_t^{(\ell)} = \frac{1}{N} \sum_{i=1}^{N} \hat{z}_{t-i} \cdot \hat{z}_{t-i-\ell},
\quad \text{where} \; N = W - \ell.
\end{equation}

A high value of $r_t^{(\ell)}$ indicates strong self-similarity after step period $\ell$, suggesting a return to previously visited semantic states and the emergence of local periodicity in the model’s reasoning path.

At each step $t$, we identify the most prominent periodicity by maximizing the correlation:
\begin{equation}
\label{eq:lag_select_final}
\ell_t^\star = \underset{\ell \in [1, P_{\max}]}{\arg\max} \; r_t^{(\ell)}, \quad
\rho_t = r_t^{(\ell_t^\star)}.
\end{equation}
The selected $\ell_t^\star$ thus serves as the most likely period of recurrence in the Long CoT trajectory, with $\rho_t$ quantifying the strength of this periodic pattern.

\textbf{Cyclic Detection via Hysteresis Control.}
The method described above estimates, in real time, the most likely cycle within the local window at each step. During Long CoT generation, brief phases of verification or reflection, as well as occasional randomness, may result in repeated cycle estimates across adjacent steps. To avoid overreacting to such transient similarities, we consider a cycle to be genuinely persistent only when the same periodic pattern is consistently detected across multiple consecutive steps.

To filter out spurious activations and ensure stable cycle detection, we implement a finite-state hysteresis mechanism controlled by two hyperparameters: a confidence threshold \(\rho^*
\) and a stability duration \(M\).
\begin{itemize}[leftmargin=1em]
     \item \textbf{Enter Cycle:} The system transitions to the \textsc{Cycle} state when \(\rho_t \geq \rho^*\) and \(\ell_t^\star\) remains stable for \(M\) consecutive steps.
     \item \textbf{Exit Cycle:} The system returns to the \textsc{Normal} state when \(\rho_t < \rho^*\) or \(\ell_t^\star\) deviates by more than one.
\end{itemize}

\vspace{0pt}
Please refer to Sec.~\ref{sec:exp-setup} for specific hyperparameter settings.

\subsection{Early-Exit Policy}
\label{sec:early_exit}
Inspired by the recent work in early-exit mechanisms~\cite{yang2025dynamic,muennighoff2025s1}, we introduce a policy that enables the LRMs to terminate the reasoning process once cyclic redundancy is detected by \model. Upon identifying such a cycle, the model is prompted to pause and produce an intermediate answer based on the reasoning trajectory generated so far. Concretely, we append a special end-of-thinking delimiter followed by the phrase ``\texttt{Final Answer:}'' to trigger an early exit from the reasoning phase and initiate answer generation.
The whole workflow of the proposed \model and the early exit policy is illustrated in Fig.~\ref{img:framework}.









\section{Experiments}

\begin{table}[t]
\caption{Main results of R1-Distilled 7/14/32B models on GSM8K, MATH, and AIME. We report accuracy and average token length (with reduction ratio) for comparison. \textcolor{darkgreen}{Green} texts mark the anticipated positive improvements.}
\vspace{2pt}
\label{tab:main}
\centering
\small
\setlength{\tabcolsep}{1mm}
\begin{tabular}{@{}lllllll@{}}
\toprule
\multirow{2}{*}{\textbf{Methods}} & \multicolumn{2}{c}{\textbf{GSM8K}} & \multicolumn{2}{c}{\textbf{MATH-500}} & \multicolumn{2}{c}{\textbf{AIME}} \\ \cmidrule(lr){2-3} \cmidrule(lr){4-5} \cmidrule(lr){6-7}
 & Acc(\%)$\uparrow$ & Len$\downarrow$ & Acc(\%)$\uparrow$ & Len$\downarrow$ & Acc(\%)$\uparrow$ & Len$\downarrow$ \\ 
\midrule

\multicolumn{7}{c}{\textit{DeepSeek-R1-Distill-Qwen-7B}} \\ \cmidrule{1-7}
Original & 90.78 & 1057.5 & 85.00 & 2857.1 & 50.00 & 10570.4 \\
\hdashline[1pt/2pt]
\addlinespace[2pt]
D-Prompt & 90.59\cred{(-0.92)} & 906.2\cgreen{(-14.3\%)} & 85.83\cgreen{(+0.83)} & 2726.1\cgreen{(-4.6\%)} & 46.67\cred{(-3.33)} & 10433.1\cgreen{(-1.5\%)} \\
NoThinking & 88.47\cred{(-3.04)} & 243.9\cgreen{(-76.9\%)} & 79.38\cred{(-5.62)} & 699.0\cgreen{(-75.5\%)} & 40.00\cred{(-10.0)} & 6382.7\cgreen{(-39.6\%)} \\
\hdashline[1pt/2pt]
\addlinespace[2pt]
DEER & 91.25\cgreen{(+0.53)} & 654.1\cgreen{(-38.11\%)} & 85.42\cgreen{(+0.42)} & 1518.2\cgreen{(-46.8\%)} & 36.67\cred{(-13.33)} & 9344.9\cgreen{(+11.5\%)} \\
\midrule
\rowcolor{mygray} \textbf{CoRE-Eval} & \textbf{92.11}\cgreen{(+1.52)} & 762.0\cgreen{(-27.9\%)} & \textbf{88.33}\cgreen{(+3.33)} & 1906.0\cgreen{(-33.2\%)} & \textbf{50.00}\cgreen{(+0.0)} & 9120.4\cgreen{(-13.7\%)} \\

\midrule
\multicolumn{7}{c}{\textit{DeepSeek-R1-Distill-Qwen-14B}} \\ \cmidrule{1-7}
Original & 92.27 & 597.2 & 86.46 & 2306.0 & 50.00 & 9900.1 \\

\hdashline[1pt/2pt]
\addlinespace[2pt]
D-Prompt & 92.76\cgreen{(+0.43)} & 606.6\cred{(+3.7\%)} & 86.46\cgreen{(+0.0)} & 2297.9\cgreen{(-0.3\%)} & 50.00\cgreen{(+0.0)} & 9839.5\cgreen{(-0.1\%)} \\
NoThinking & 88.41\cred{(-3.92)} & 205.7\cgreen{(-64.9\%)} & 75.83\cred{(-10.63)} & 611.8\cgreen{(-73.5\%)} & 33.33\cred{(-16.67)} & 5977.1\cgreen{(-39.6\%)} \\
\hdashline[1pt/2pt]
\addlinespace[2pt]
DEER & 89.83\cgreen{(-2.5)} & 297.8\cgreen{(-50\%)} & 86.66\cgreen{(+0.2s)} & 1632.1\cgreen{(-29.2\%)} & 50.00\cgreen{(+0.0)} & 6658.4\cgreen{(-32.4\%)} \\
\midrule
\rowcolor{mygray} \textbf{CoRE-Eval} & 90.78\cred{(-1.52)} & 419.5\cgreen{(-28.1\%)} & \textbf{87.29}\cgreen{(+0.83)} & 1827.0\cgreen{(-20.8\%)} & \textbf{53.33}\cgreen{(+3.33)} & 8432.1\cgreen{(-14.8\%)} \\

\midrule
\multicolumn{7}{c}{\textit{DeepSeek-R1-Distill-Qwen-32B}} \\ \cmidrule{1-7}
Original & 95.82 & 717.2 & 87.08 & 2356.9 & 60.00 & 9605.1 \\
\hdashline[1pt/2pt]
\addlinespace[2pt]
D-Prompt & 95.82\cgreen{(+0.0)} & 662.7\cgreen{(-7.6\%)} & 86.66\cred{(-0.42)} & 2197.9\cgreen{(-6.8\%)} & 56.67\cred{(-3.33)} & 9347.8\cgreen{(-0.3\%)} \\
NoThinking & 93.86\cred{(-1.96)} & 228.1\cgreen{(-68.2\%)} & 80.00\cred{(-7.08)} & 659.1\cgreen{(-72.0\%)} & 50.00\cred{(-10.0)}& 5612.9\cgreen{(-41.5\%)} \\
\hdashline[1pt/2pt]
\addlinespace[2pt]
DEER & 95.28\cred{(-0.54)} & 433.2\cgreen{(-39.5\%)} & 85.63\cred{(-1.46)} & 1303.1\cgreen{(-44.7\%)} & 60\cgreen{(+0.0)} & 7677.2\cgreen{(-19.8\%)} \\
\midrule
\rowcolor{mygray} \textbf{CoRE-Eval} & \textbf{96.45}\cgreen{(+0.63)} & 563.1\cgreen{(-21.5\%)} & \textbf{90.60}\cgreen{(+0.8)} & 1511.1\cgreen{(-35.9\%)} & \textbf{70.00}\cgreen{(+10.00)} & 7235\cgreen{(-24.6\%)} \\

\bottomrule
\end{tabular}
\vspace{-10pt}
\end{table}

\subsection{Experimental Setup}
\label{sec:exp-setup}

\textbf{Datasets.}
We primarily focus on mathematical reasoning tasks, as their answers are automatically verifiable, enabling scalable and objective evaluation. To comprehensively assess model performance across diverse reasoning capabilities, we adopt three representative benchmarks widely used in the literature: \textsc{GSM8K}~\cite{cobbe2021training}, \textsc{MATH-500}~\cite{hendrycks2021measuring}, and \textsc{AIME 2024}. \textsc{GSM8K} is a crowd-sourced dataset of grade-school math word problems that require multi-step arithmetic reasoning. \textsc{MATH-500} is a challenging subset of the MATH dataset, curated by OpenAI. \textsc{AIME 2024} consists of 30 problems selected from the 2024 American Invitational Mathematics Examination.

To analyze the geometric properties of \modelname~(Sec.~\ref{sec:state_transition_diag}), we conduct trajectory visualizations on a randomly sampled subset of 50 examples from \textsc{GSM8K} dataset and 20 from \textsc{MATH-500} dataset. These samples are used solely for qualitative analysis and are excluded from the evaluation metrics reported in Sec.~\ref{sec:exp-results}.

\textbf{Reasoning models.}
We employ the open-source DeepSeek-R1-Distill family of models~\citep{guo2025deepseek}, including R1-Distill-Qwen-7B, R1-Distill-Qwen-14B, and R1-Distill-Qwen-32B. These models are obtained via supervised fine-tuning on reasoning data generated by the original DeepSeek-R1 model, providing a consistent and scalable foundation for evaluating self-evaluation mechanisms in long CoT generation.

\textbf{Baselines.}
We compare \textbf{CoRE-Eval} with three representative methods: two prompt-based and one confidence-based. \textbf{D-Prompt} encourages the model to adjust its reasoning length based on self-assessed task difficulty, using instructions such as ``\texttt{Please think quickly if you find it easy}'' to prompt concise answers for easier problems. \textbf{NoThinking}~\cite{ma2025reasoning} bypasses the reasoning process entirely by prompting the model to directly output answers without intermediate steps, regardless of problem difficulty. \textbf{DEER}~\cite{yang2025dynamic} leverages semantic markers (e.g., ``\texttt{Wait}'' tokens) to detect reasoning transitions and halts generation when the model exhibits high confidence in a trial answer. As the official implementation of DEER is not publicly available, we reimplement it within our experimental framework for fair comparison.

\textbf{Metrics.}
We evaluate model performance using two primary metrics. \textit{Accuracy} measures the proportion of examples in which the model’s final answer exactly matches the ground-truth label. Given that all benchmark datasets provide deterministic and verifiable solutions, this metric directly reflects end-task correctness. \textit{Average Output Length} is defined as the mean number of tokens generated in the model’s reasoning trace and serves as a proxy for reasoning verbosity and cognitive load. This metric is particularly relevant for assessing the efficiency of early-exit mechanisms and their impact on reasoning conciseness.

\textbf{Implementation details.}
We perform all evaluation experiments in a Zero-shot CoT setting, utilizing the prompt: ``\texttt{Please reason step by step, and put your final answer within \textbackslash{boxed$\{\}$}.}'' For decoding, we employ greedy search with a single sample to evaluate correctness. Given that ground-truth answers are well-structured numerical values or options, mathematical equivalence is verified directly via rule-based evaluations. To ensure complete problem-solving attempts are captured, the maximum new token length (``\textit{max\_new\_tokens}'') is set to 16384 for AIME2024 and 8192 for other datasets. Hyperparameter settings include $\rho_{\text{*}}\!=\!0.7$, $P_{max}\!=\!8$, $W\!=32$. All experiments are conducted with PyTorch 2.6.0 on 2$\times$ NVIDIA GeForce RTX V100 GPUs.

\subsection{Experimental Results}
\label{sec:exp-results}

\textbf{Main Results.}  
As shown in Tab.~\ref{tab:main}, \textbf{CoRE-Eval} consistently achieves significant improvements across all evaluated reasoning models and datasets~\footnote{We use selected samples from \textsc{MATH-500} and \textsc{GSM8K} for qualitative analysis. The evaluation set contains 1,269 and 480 problems, respectively.}. Specifically, \textbf{CoRE-Eval} enhances accuracy by an average of 3.58\% while simultaneously reducing the length of CoT sequences by 13.7\% to 33.2\%. These gains are observed across diverse benchmarks, including \textsc{GSM8K}, \textsc{MATH-500}, and \textsc{AIME}, and across model sizes ranging from 7B to 32B.

Notably, when applied to the \textsc{AIME} benchmark using the 32B model, \textbf{CoRE-Eval} achieves an impressive 70\% accuracy, representing a remarkable absolute gain of 10\% over the baseline, while reducing the reasoning length by 24.6\%. This result highlights \textbf{CoRE-Eval}'s ability to enhance reasoning precision and efficiency even on highly challenging mathematical reasoning tasks.

\begin{wrapfigure}{r}{0.55\textwidth}
\vspace{-0.25in}
  \centering
   \includegraphics[width=0.27\textwidth]{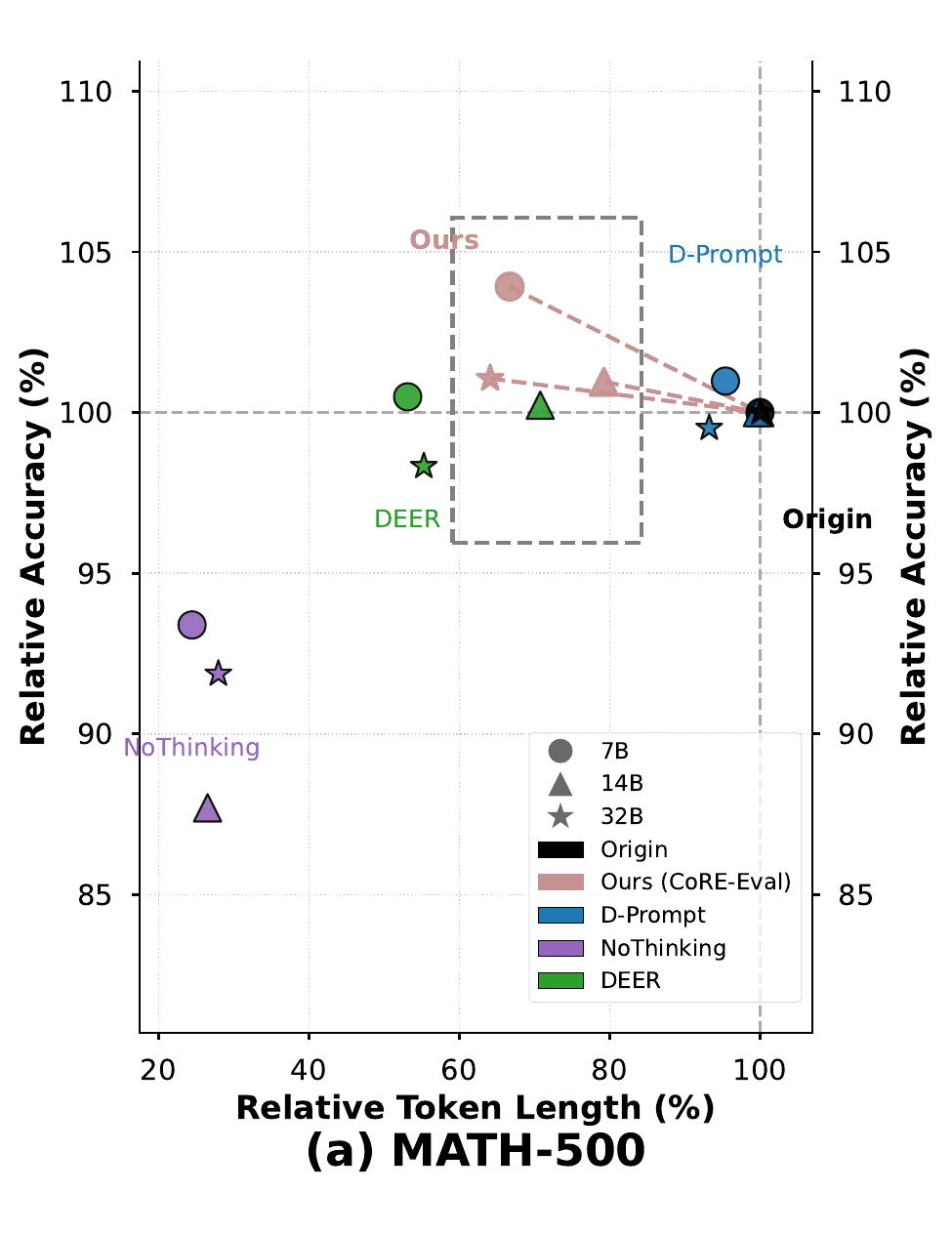}
   \includegraphics[width=0.27\textwidth]{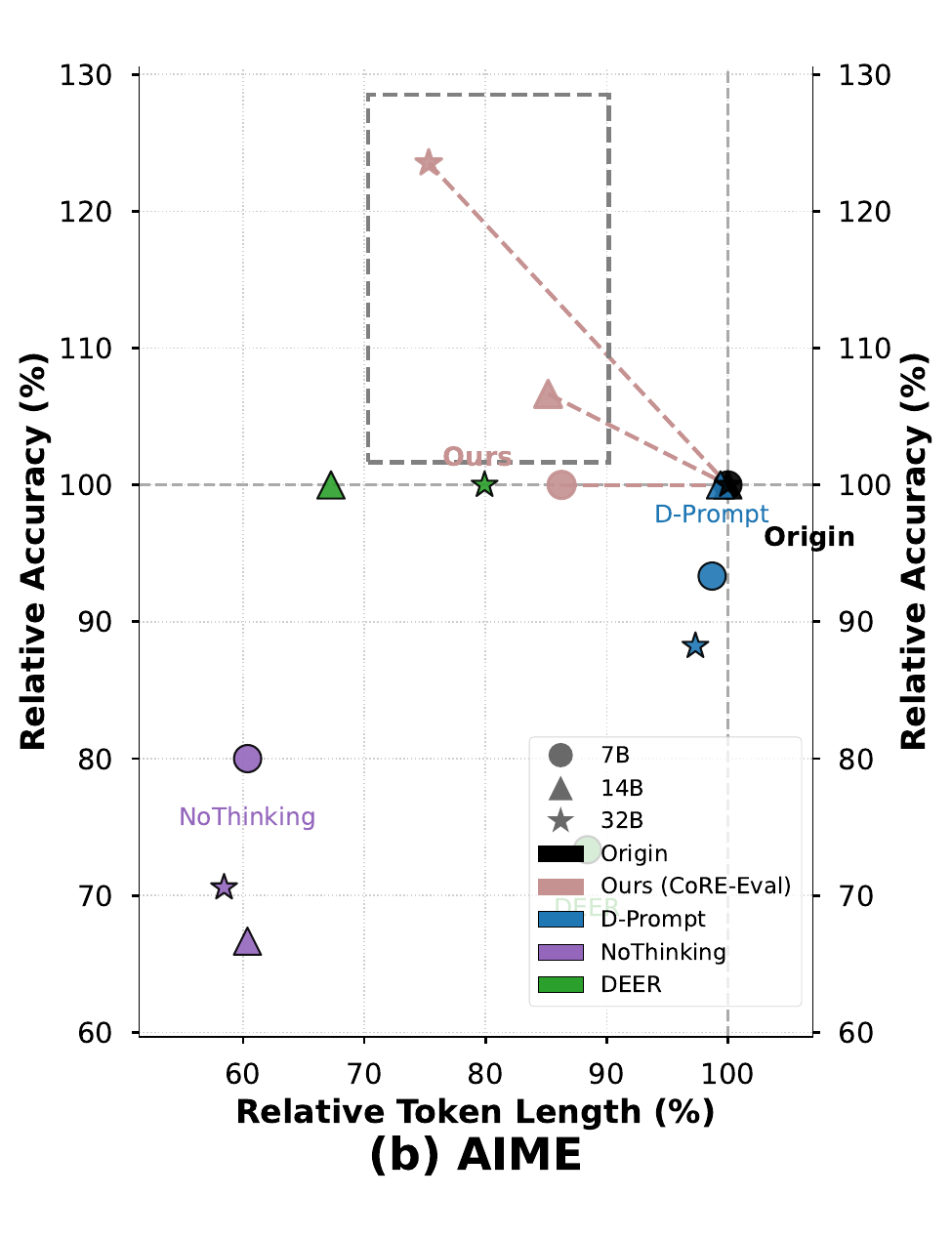}
  \vspace{-0.25in}
  \caption{Relative accuracy vs. token length reduction on (a) \textsc{MATH-500} and (b) \textsc{AIME}.}
  \label{img:aime_math}
\vspace{-0.2in}
\end{wrapfigure}

In contrast, baseline approaches such as D-Prompt, NoThinking, and DEER are effective in shortening reasoning sequences but often at the cost of decreased accuracy. This suggests that while these methods reduce verbosity, they may compromise reasoning completeness or correctness. By comparison, \textbf{CoRE-Eval} achieves a better balance between efficiency and accuracy, underscoring its robustness as a self-evaluation mechanism for mitigating overthinking in LRMs.

\textbf{Fine-Grained Analysis.}
We further conduct a fine-grained analysis to compare \textbf{CoRE-Eval} with baseline methods, focusing on the trade-off between reasoning efficiency and accuracy.

As shown in Fig.~\ref{img:aime_math}, \textbf{CoRE-Eval} demonstrates a unique advantage by achieving higher relative accuracy while effectively reducing reasoning steps. This highlights its metacognitive capability to identify and avoid redundant reasoning. In contrast, \textbf{D-Prompt} relies on heuristic instructions to control reasoning length, but struggles to mitigate cognitive overload in LRMs. \textbf{NoThinking} aggressively suppresses reasoning steps, resulting in the lowest token lengths across settings; however, this comes at the cost of significant accuracy degradation, as models that are not allowed to reason cannot produce reliable answers. Similarly, the confidence-based \textbf{DEER} approach is prone to premature termination due to overconfidence in the early stages of reasoning, particularly in larger models, leading to reduced token usage but also decreased accuracy.

\begin{wrapfigure}{r}{0.55\textwidth}
\vspace{-0.25in}
  \centering
   \includegraphics[width=0.53\textwidth]{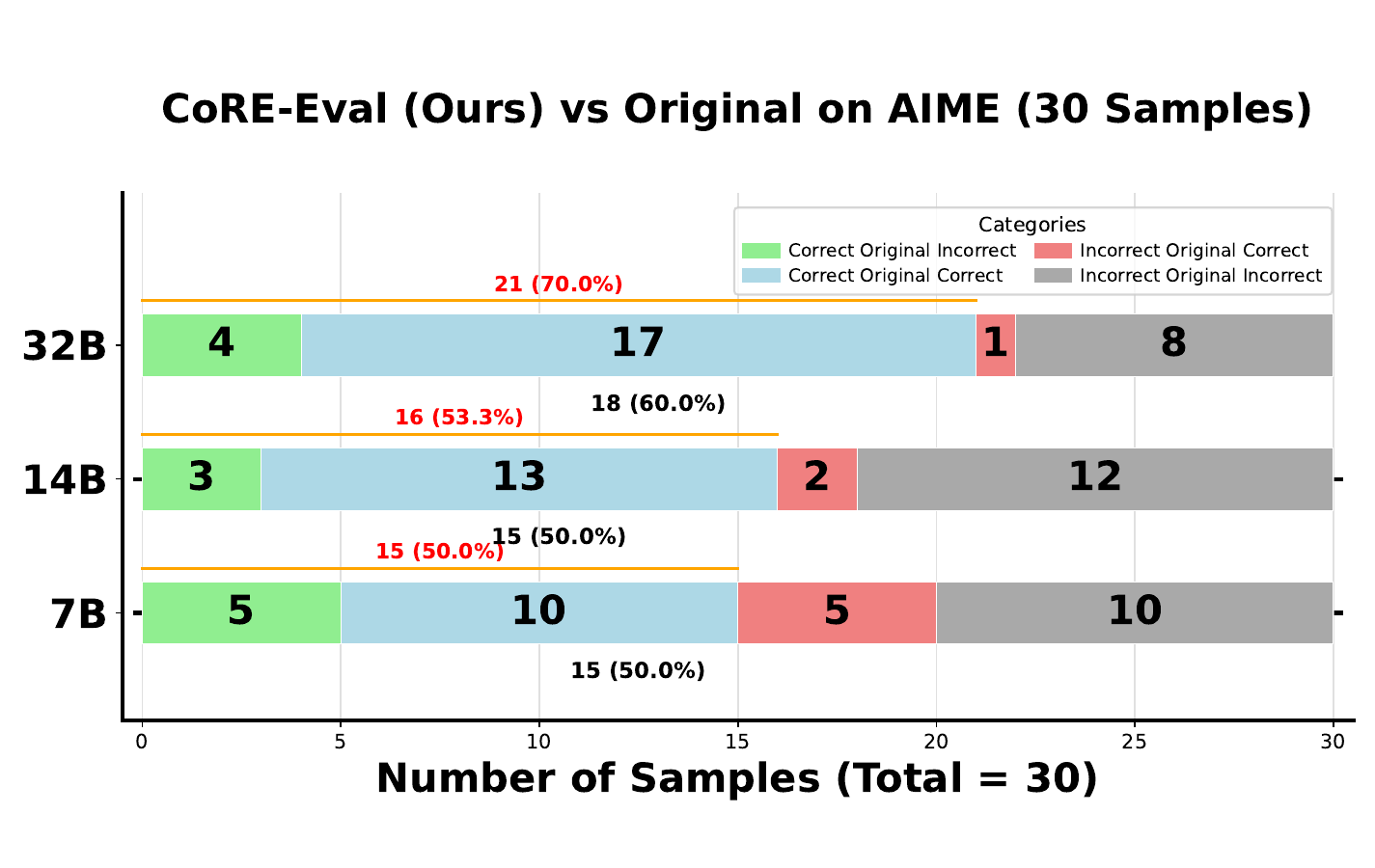}
   \vspace{-0.2in}
  \caption{\textbf{CoRE-Eval} vs. Original (i.e., on \textsc{AIME}).}
  \label{img:core_vs_origin}
\vspace{-0.10in}
\end{wrapfigure}

In addition, the breakdown in Fig.~\ref{img:core_vs_origin} shows that \textbf{CoRE-Eval} consistently corrects more answers through early exits (green bars) than it mistakenly alters (red bars), especially as model size increases. This suggests that our method is more effective when applied to more capable models, leveraging their stronger reasoning abilities to perform more reliable self-evaluation. These observations collectively reinforce the scalability and robustness of \textbf{CoRE-Eval} in enhancing LRM efficiency without compromising correctness.



\subsection{Ablation Study}
\begin{wrapfigure}{r}{0.55\textwidth}
\vspace{-0.25in}
  \centering
   \includegraphics[width=0.27\textwidth]{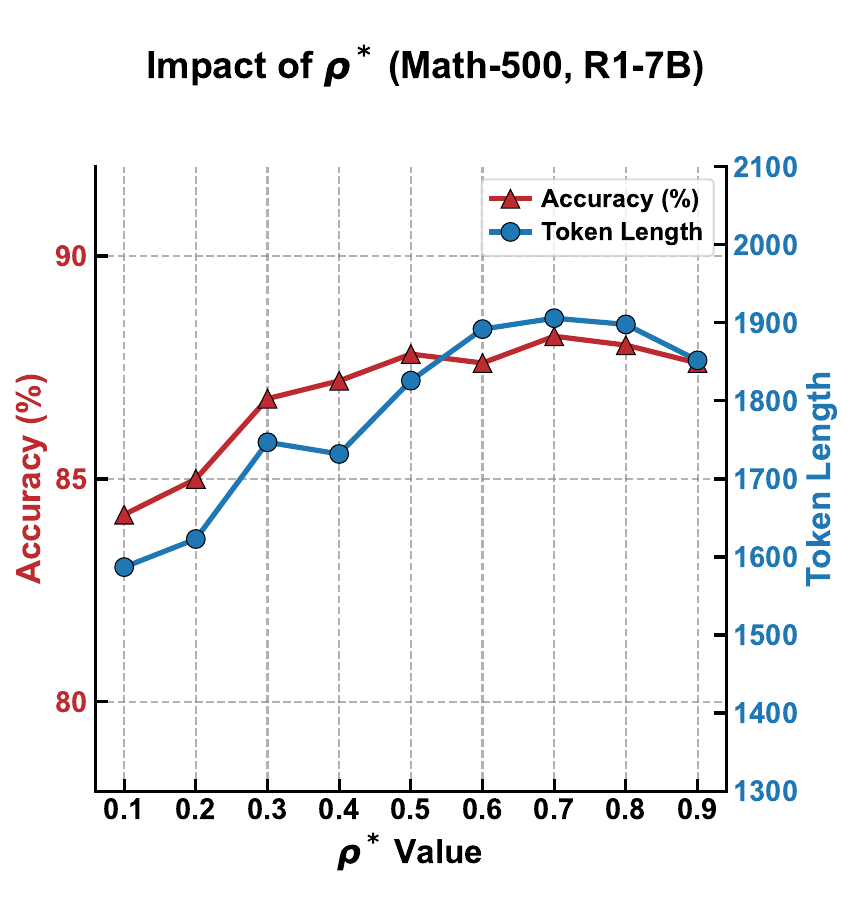}
   \includegraphics[width=0.27\textwidth]{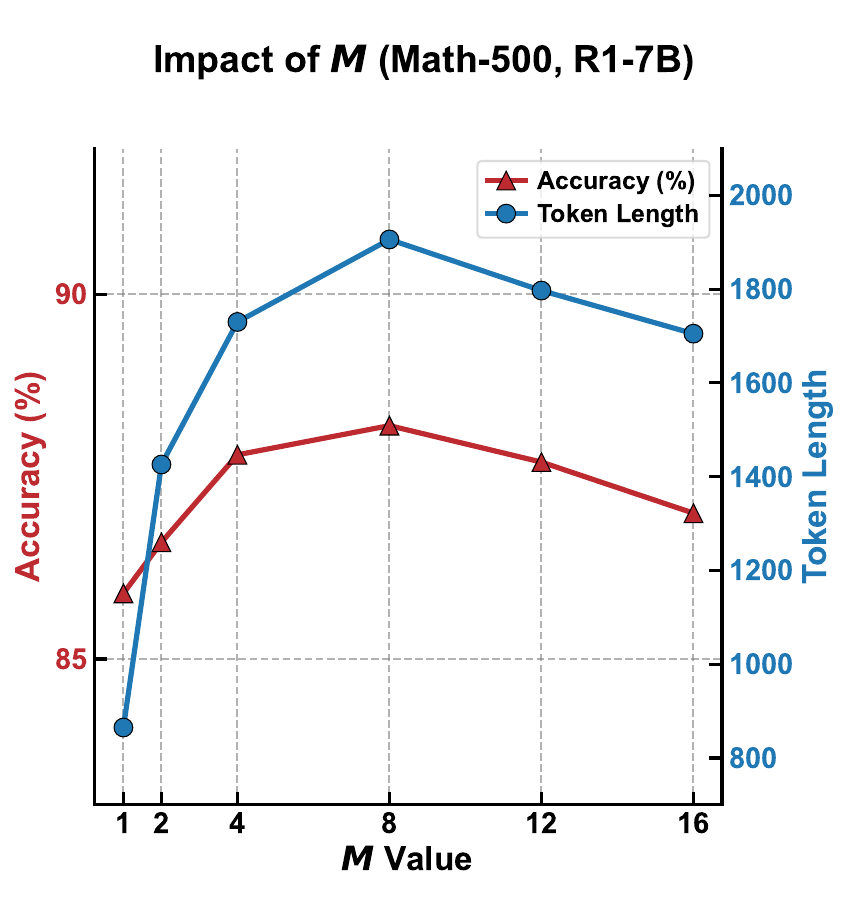}
  \vspace{-0.20in}
  \caption{Ablation Study.}
  \label{img:ablation}
\vspace{-0.10in}
\end{wrapfigure}

\textbf{Impact of $\rho^*$.}  
The threshold $\rho^*$ controls the sensitivity of cycle detection by specifying the minimum correlation strength required to recognize a cyclic pattern. As shown in Fig.~\ref{img:ablation}, low values (e.g., 0.1--0.3) cause premature detection of unstable patterns, leading to overly aggressive early exits and degraded accuracy despite shorter reasoning traces. In contrast, higher thresholds (0.7--0.8) allow the model to trigger early exits only upon identifying robust and confident cycles, striking a better balance between reasoning efficiency and accuracy. The best result is observed at $\rho^*=0.7$, yielding an accuracy of 88.20\% with an average token length of 1906. Beyond this point, the model becomes overly conservative, slightly reducing both accuracy and efficiency.

\textbf{Impact of $M$.}  
We also analyze the effect of the stability duration $M$, which determines how many consecutive steps must confirm a cycle before early exit is triggered. As shown in Fig.~\ref{img:ablation}, increasing $M$ from 1 to 8 steadily improves performance, as it helps the model avoid reacting to transient oscillations. The best trade-off is achieved at $M=8$, with an accuracy of 88.20\% and a reasoning length of 1906 tokens. Larger values of $M$ (12 or 16) make the model more hesitant to exit, slightly diminishing both accuracy and efficiency. This confirms that moderate stability requirements enable \textbf{CoRE-Eval} to robustly filter false alarms while maintaining its metacognitive advantage.

\section{Conclusion}
\label{sec:conclusion}

We present \textbf{CoRE-Eval}, a label-free self-evaluation framework that enhances the metacognitive abilities of large reasoning models (LRMs) by analyzing their latent reasoning trajectories. Through the proposed Chain-of-Reasoning Embedding (CoRE), we capture interpretable geometric signals to detect cyclic redundancy and enable early termination of inefficient reasoning. Experiments on mathematical reasoning benchmarks show that \textbf{CoRE-Eval} reduces reasoning length while improving accuracy across model scales, achieving 70.0\% accuracy on \textsc{AIME 2024} with a 32B model. Our work provides an efficient and scalable solution to mitigating overthinking in LRMs, opening avenues for integrating introspective reasoning control into broader domains.

{\small
\bibliographystyle{plain}
\bibliography{neurips_2025}
}
\newpage
\section*{Appendix}
\label{sec.appendix}
\subsection{Limitations}
\label{subsec.lim}

While \textbf{CoRE-Eval} offers an effective label-free self-evaluation framework, it has two primary limitations. First, our method requires access to step-level hidden states, restricting its applicability to white-box models and preventing direct deployment on closed-source systems such as OpenAI's O1. Nevertheless, this design choice aligns with our focus on interpretability and fosters deeper understanding of model reasoning dynamics. Second, the step-wise correlation computation introduces latency overhead, particularly on simpler tasks like \textsc{GSM8K}, where overthinking is less prevalent and efficiency gains are limited. Future work may explore adaptive or intermittent detection strategies to alleviate this issue while preserving reasoning efficiency.

\subsection{Broader Impacts}
\label{subsec:impacts}
Our work presents \textbf{CoRE-Eval}, a label-free self-evaluation framework for large reasoning models (LRMs) that enhances reasoning efficiency by mitigating overthinking through geometric trajectory analysis. The positive societal impacts of our approach lie in its potential to reduce computational overhead and environmental costs associated with inference in large-scale models, enabling more energy-efficient and accessible deployment of advanced reasoning systems. Furthermore, by improving the metacognitive capabilities of LRMs, our method may contribute to more reliable and interpretable AI systems in critical domains such as mathematics education, scientific discovery, and decision support. 

\subsection{Hyperparameter Settings.}  
\label{subsec.Hyperparameter}
Through qualitative observations of \textbf{CoRE} trajectories on samples from \textsc{MATH-500} and \textsc{GSM8K}, we found that the dominant periodic patterns typically occur within cycles of length 2 to 6, with no cycles exceeding 8 steps. Based on this empirical insight, we set the maximum period $P_{\max}=8$. To ensure the sliding window can capture at least two full cycles of the longest possible period, we fix the window size as $W=4P_{\max}=32$ throughout our experiments.

\subsection{Implementation Details}
\begin{figure*}[h]
    \centering
    \includegraphics[width = 1.0\textwidth]{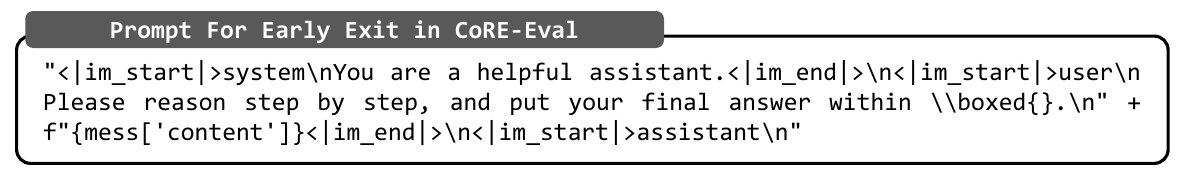}
    \caption{Prompt Template for \textbf{CoRE-Eval}}
    \label{fig:prompt_inference}
\end{figure*}

\textbf{Prompts.}
Fig.~\ref{fig:prompt_inference} presents the prompt used to elicit reasoning traces from all LRMs. For the Qwen models, we slightly modified the original prompt. While we observed a minor performance drop on the benchmark, it remains within an acceptable range. To ensure experimental consistency, we used the same prompt when extracting reasoning embeddings for each reasoning step in \textbf{CoRE}.

\begin{figure*}[h]
    \centering
    \vspace{-10pt}
    \includegraphics[width = 1.0\textwidth]{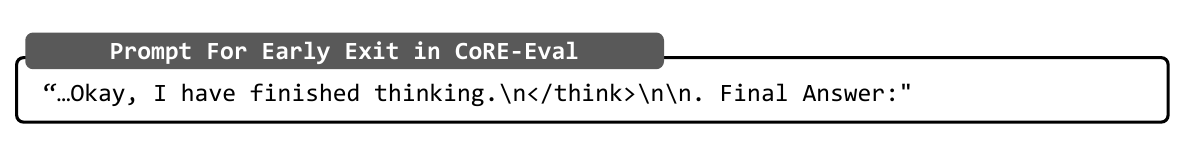}
    \caption{Prompt For Early Exit in \textbf{CoRE-Eval}}
    \label{fig:prompt_early}
\end{figure*}

Fig.~\ref{fig:prompt_early} shows the prompt that enables the LRMs to terminate the reasoning process once cyclic redundancy is detected by \textbf{CoRE-Eval}.

\begin{figure*}[h]
    \centering
    \includegraphics[width = 1.0\textwidth]{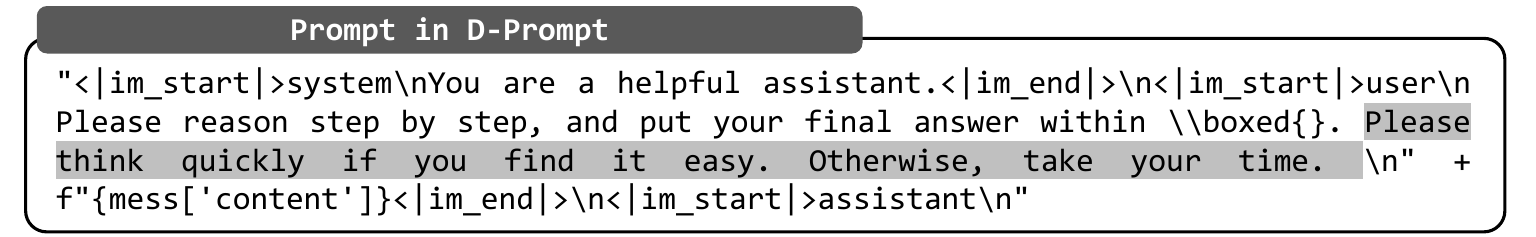}
    \caption{Prompt in \textbf{D-Prompt}}
    \vspace{-5pt}
    \label{fig:prompt_D}
\end{figure*}

What's more, we compare \textbf{CoRE-Eval} with two prompt-based approaches. Here we also provide the detailed prompts usage in our experiments. Fig.~\ref{fig:prompt_D} is the prompt used to elicit reasoning traces from all LRMs for the baseline \textbf{D-Prompt}. 

\begin{figure*}[h]
    \centering
    \includegraphics[width = 1.0\textwidth]{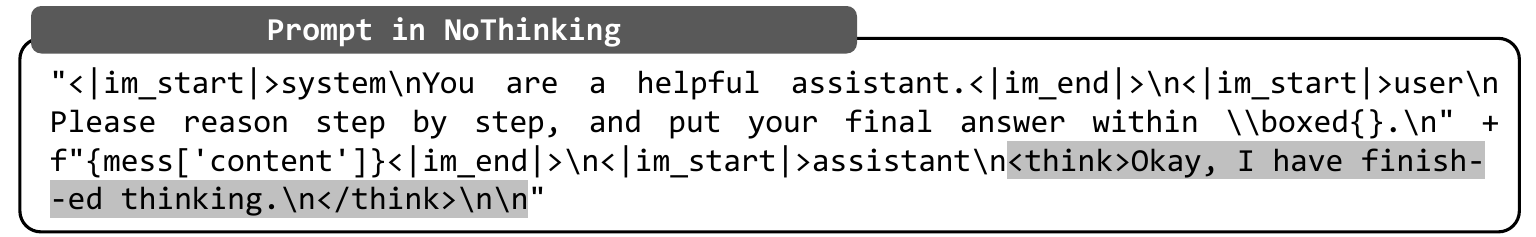}
    \caption{Prompt in \textbf{NoThinking}}
    \label{fig:prompt_N}
\end{figure*}

Fig.~\ref{fig:prompt_N} is the setting of baseline \textbf{NoThinking}.

\textbf{Dataset Details.}  
In our main experiments, we evaluate model performance on three widely adopted mathematical reasoning benchmarks:

\textbf{\textsc{GSM8K}} is a crowd-sourced dataset of grade-school level math word problems that require multi-step arithmetic reasoning. It serves as a standard benchmark for evaluating mathematical language understanding in large language models.

\textbf{\textsc{MATH-500}} is a challenging subset of the MATH dataset, containing 500 competition-level problems spanning high school domains such as prealgebra, algebra, geometry, number theory, and combinatorics. We follow prior work and adopt the 500-problem split curated by OpenAI to ensure consistency in evaluation.

\textbf{\textsc{AIME 2024}} consists of 30 problems selected from the 2024 American Invitational Mathematics Examination. This prestigious competition assesses advanced mathematical reasoning skills across a wide range of topics, including algebra, counting, geometry, number theory, and probability.

Beyond the main benchmarks, we also evaluate our method on more advanced mathematical and programming tasks to assess generalization (See Appx.~\ref{subsec.exper}):

\textbf{\textsc{AIMO Validation AMC}} includes 83 problems drawn from AMC12 2022 and 2023, covering topics such as algebra, geometry, number theory, and combinatorics. This dataset was used for internal validation during the AIMO Progress Prize competition. To reduce overlap with the MATH training set, it was restricted to post-2021 questions. Original AMC12 problems are multiple-choice, but to align with the AIMO evaluation format, we manually reformulated solvable questions to require an integer answer. Problems that could not be reformulated were excluded.

\textbf{\textsc{HumanEval}} is a benchmark introduced by OpenAI for evaluating code generation. It consists of 164 hand-curated Python programming tasks covering algorithmic problem solving. Each example includes a function signature, a natural language docstring, a reference solution, and test cases for correctness verification.

\subsection{Additional Experimental Results}
\label{subsec.exper}

Beyond the main mathematical reasoning benchmarks, we further evaluate \textbf{CoRE-Eval} on two additional high-difficulty tasks to assess its generalization capabilities across domains: \textsc{AIMO Validation AMC}, a competition-style mathematical dataset, and \textsc{HumanEval}, a code generation benchmark. These datasets highlight the applicability of our method in both symbolic and algorithmic reasoning settings. We directly reuse the hyperparameter settings from previous experiments for consistency (see Sec.~\ref{sec:exp-setup} and Sec.~\ref{subsec.Hyperparameter}).

\begin{table}[t]
\caption{Results of R1-Distill-Qwen-7B on \textsc{AIMO Validation AMC} and \textsc{HumanEval}. We report accuracy for AMC and pass@1 for HumanEval, along with average token length.}
\label{tab:main-aimo-humaneval}
\vspace{2pt}
\centering
\small
\setlength{\tabcolsep}{4mm}
\renewcommand{\arraystretch}{1.2}
\begin{tabular}{@{}lcccc@{}}
\toprule
\textbf{Methods} 
& \multicolumn{2}{c}{\textsc{AIMO Validation AMC}} 
& \multicolumn{2}{c}{\textsc{HumanEval}} \\
\cmidrule(lr){2-3} \cmidrule(lr){4-5}
& Accuracy (\%) $\uparrow$ & Len$\downarrow$ 
& Pass@1 (\%) $\uparrow$ & Len$\downarrow$ \\
\midrule
\multicolumn{5}{c}{\textit{\textbf{DeepSeek-R1-Distill-Qwen-7B}}} \\
\cmidrule{1-5}
Original & 59.03 & 5059 & 48.78 & 2992 \\
\hdashline[1pt/2pt]
\rowcolor{mygray} \textbf{CoRE-Eval} & \textbf{66.26}\cgreen{(+7.23)} & \textbf{4145}\cgreen{(-18.07\%)} & \textbf{50.61}\cgreen{(+1.83)} & \textbf{2144}\cgreen{(-28.34\%)} \\
\bottomrule
\end{tabular}
\vspace{-5pt}
\end{table}
As shown in Table~\ref{tab:main-aimo-humaneval}, \textbf{CoRE-Eval} delivers consistent improvements over the Original baseline across both tasks. On \textsc{AIMO Validation AMC}, our method achieves a relative gain of +7.23\% in accuracy while reducing average reasoning length by 18.07\%. This indicates that CoRE-Eval effectively mitigates overthinking even in highly challenging math problems derived from AMC competitions, demonstrating strong self-evaluation under limited supervision.

On \textsc{HumanEval}, which involves Python function synthesis from natural language prompts, \textbf{CoRE-Eval} achieves a +1.83\% gain in pass@1 while also reducing the average token length by 28.34\%. This highlights the method's ability to generalize beyond symbolic math and into structured algorithmic generation tasks, reducing verbosity without sacrificing correctness.

These results confirm that \textbf{CoRE-Eval} is a broadly applicable, domain-agnostic self-evaluation mechanism capable of enhancing both efficiency and correctness across diverse reasoning domains.

\subsection{Case Study}
\begin{figure*}[htbp]
    \centering
    \vspace{-10pt}
    \includegraphics[width = 1.0\textwidth]{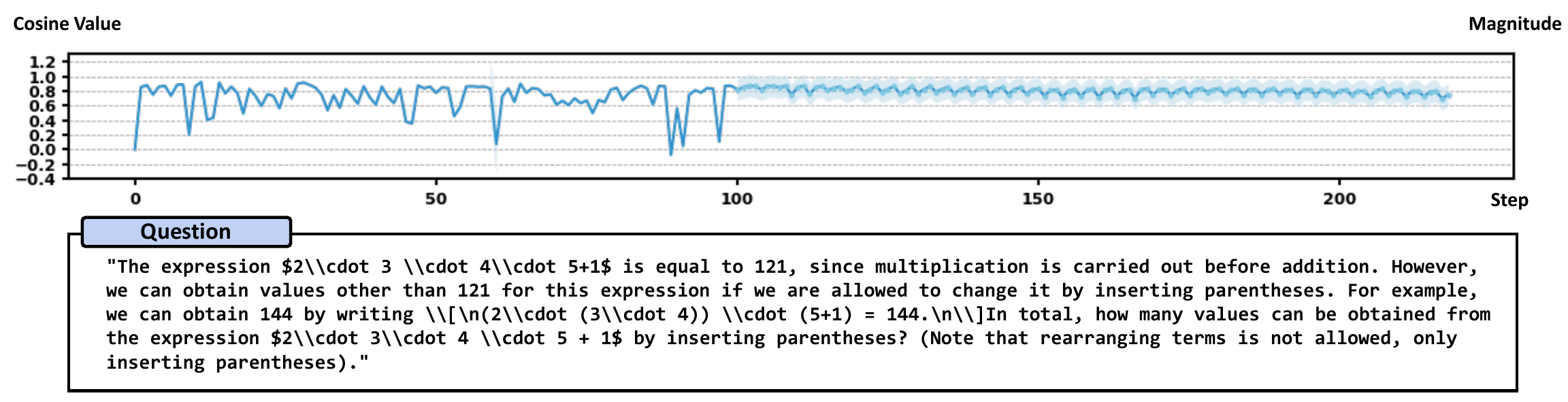}
    \caption{\textbf{CoRE-Eval} Visualization Example in \textsc{Math-500}.}
    \vspace{-10pt}
    \label{fig:case_1}
\end{figure*}

We first show more \textbf{CoRE} Visualization cases to investigate the influence of \textbf{CoRE}'s geometric features, As shown in Fig.~\ref{fig:case_1} and Fig.~\ref{fig:case_2}
\vspace{-20pt}
\begin{figure*}[htbp]
    \vspace{-20pt}
    \centering
    \includegraphics[width = 1.0\textwidth]{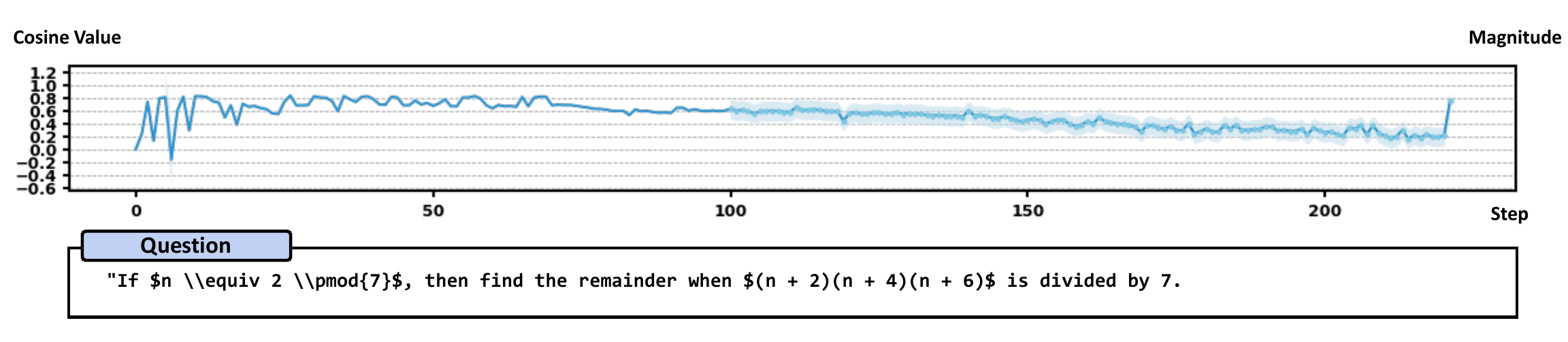}
    \caption{\textbf{CoRE-Eval} Visualization Example in \textsc{Math-500}}
    \vspace{-20pt}
    \label{fig:case_2}
\end{figure*}
\begin{figure*}[htbp]
    \centering
    \vspace{-10pt}
    \includegraphics[width = 1.0\textwidth]{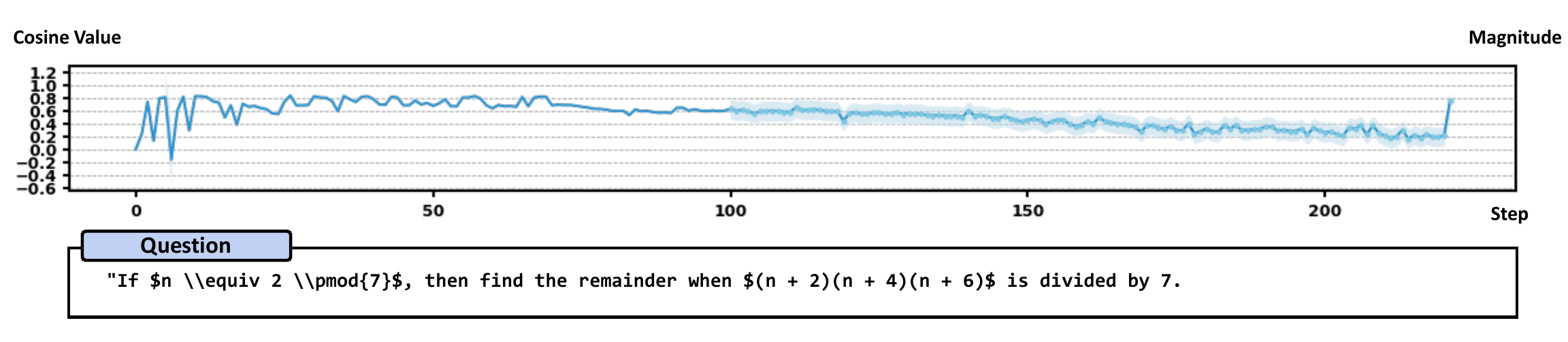}
    \caption{\textbf{CoRE-Eval} Visualization Example in \textsc{GSM8K}}
    \vspace{-20pt}
    \label{fig:case_2}
\end{figure*}

\begin{figure*}[htbp]
    \centering
    \includegraphics[width = 1.0\textwidth]{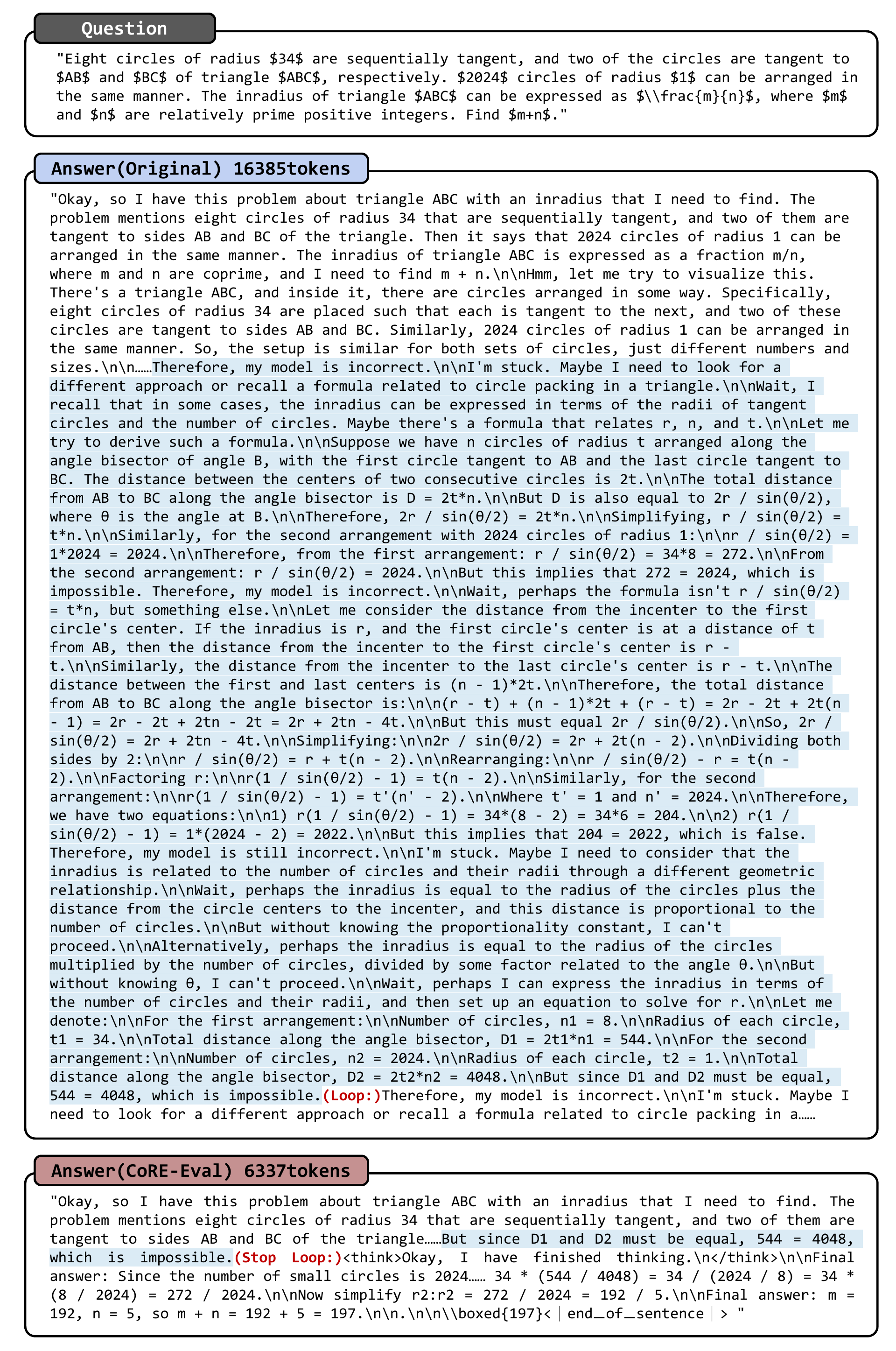}
    \caption{An example demonstrating \textbf{CoRE-Eval} Metacognition capability and concise response.}
    \label{fig:case_study_easy_easy}
\end{figure*}

Here we give concrete case that in \textsc{AIME 2024} in Fig.~\ref{fig:case_study_easy_easy}.

\end{document}